\documentclass[11pt,a4paper]{article}
\usepackage[T1]{fontenc}
\usepackage[utf8]{inputenc}
\usepackage[english]{babel}
\usepackage{lmodern}
\usepackage{amsmath,amssymb,amsthm,mathtools}
\usepackage{booktabs,graphicx,microtype,enumitem,listings}
\usepackage[margin=2.5cm]{geometry}
\usepackage[dvipsnames]{xcolor}
\usepackage{caption}
\usepackage[colorlinks=true,linkcolor=NavyBlue,citecolor=NavyBlue,
            urlcolor=NavyBlue]{hyperref}

\IfFileExists{numbers_final.tex}{
\providecommand{\FIaPinn}{3.84\times 10^{-3}}

\providecommand{\FIagainTau}{2.2}
\providecommand{\FIagainEB}{4.0}

\providecommand{\nparams}{12\,737}
\providecommand{\FIIaPinn}{1.19\times 10^{-2}}

\providecommand{\FIIaPinnex}{1.21\times 10^{0}}
\providecommand{\FIIaPinntau}{8.00\times 10^{-2}}

\providecommand{\FIIaEbnetex}{6.81\times 10^{-3}}

\providecommand{\FIIagainCoords}{40.8}
\providecommand{\FIIaexRatio}{178}

\providecommand{\FIIbgainCoords}{15.5}

\providecommand{\FIIIagainPhase}{5.7}

}{}
\IfFileExists{numbers.tex}{

\providecommand{\nparams}{12\,737}
\providecommand{\WmeanPinn}{3.64\times 10^{-2}}

\providecommand{\WmeanPinntau}{1.99\times 10^{-2}}

\providecommand{\WmeanEbc}{1.05\times 10^{-2}}

\providecommand{\WmeanEbnet}{1.04\times 10^{-2}}

\providecommand{\extrapPinn}{1.71\times 10^{-1}}
\providecommand{\extrapPinntau}{4.61\times 10^{-2}}
\providecommand{\extrapEbnet}{3.91\times 10^{-3}}

}{}

\providecommand{\rmd}{\mathrm{d}}
\providecommand{\rmi}{\mathrm{i}}
\providecommand{\rme}{\mathrm{e}}
\providecommand{\etal}{\textit{et al}}
\newcommand{\EB}{\textsc{EB-Net}}

\theoremstyle{plain}
\newtheorem{proposition}{Proposition}
\newtheorem{lemma}[proposition]{Lemma}
\theoremstyle{remark}
\newtheorem{remark}[proposition]{Remark}

\lstdefinestyle{py}{language=Python, basicstyle=\ttfamily\scriptsize,
  keywordstyle=\color{NavyBlue}\bfseries,
  commentstyle=\color{OliveGreen}\itshape,
  numbers=left, numberstyle=\tiny\color{gray}, numbersep=6pt,
  frame=leftline, framesep=5pt, rulecolor=\color{gray!40},
  breaklines=true, showstringspaces=false, columns=fullflexible,
  xleftmargin=11pt}
\title{\bfseries Ermakov--Based embeddings as a parameter-free\\
layer for physics-informed neural networks}
\author{
Roberto Bernal-Jaquez\thanks{Corresponding author: rbernal@cua.uam.mx}\\[2pt]
\normalsize Departamento de Matem\'aticas Aplicadas y Sistemas,\\
\normalsize Universidad Aut\'onoma Metropolitana--Cuajimalpa, Mexico City, Mexico
\and
Roberto Romero Huerta\\[2pt]
\normalsize Posgrado en Ciencias Naturales e Ingenier\'ia,\\
\normalsize Universidad Aut\'onoma Metropolitana--Cuajimalpa, Mexico City, Mexico
\and
Alexander Schaum\\[2pt]
\normalsize Department for Process Analytics,\\
\normalsize University of Hohenheim, Stuttgart, Germany
}
\date{}

\begin{document}
\maketitle

\begin{abstract}
Evolution equations with time-dependent coefficients force a neural solver to spend capacity on a scale that the equation already determines. We show that a single classical object, the Ermakov-Pinney scale function $\rho$, can be written into the input layer of a physics-informed neural network as a parameter-free embedding. One construction covers three problem families through the two signs of the Pinney constant $c$: linear diffusion ($c=-1$, $\omega\equiv0$), diffusion in a time-dependent confining trap ($c=-1$, $\omega\neq0$), and the quantum oscillator with time-dependent frequency ($c=+1$). An exact moment identity gives $\rho$ in closed form from the initial data alone, so no auxiliary network or knowledge of the solution is needed. Our central finding is comparative: each family requires a different component of the construction and has a different honest control. In family I, the diffusive-time reparametrisation alone reproduces most of the gain, so that family is a poor test of the idea and shows why an ablation is mandatory. In family II no time reparametrisation exists, the control degrades accuracy, and the amplitude prefactor becomes essential (40.8x). In family III the adiabatic approximation also fails, and the quadratic phase factor is what matters (5.7x). The method does not help when the reduced solution does not converge to a fixed point: its extrapolation advantage then disappears. The reduction itself is the classical scaling-variable change of the self-similar-asymptotics literature; what is new is the moment identity that makes it architectural for time-dependent coefficients, and the comparative anatomy above.
\end{abstract}


%
%

\section{Introduction}

Physics-informed neural networks (PINNs) \cite{raissi2019} struggle when a
solution develops multiple scales or when coefficients vary rapidly
\cite{krishnapriyan2021,wang2023expert}. Supplying the network with structure
that the equation already possesses is a natural remedy, and symmetry has been
injected into neural PDE solvers as data augmentation \cite{brandstetter2022},
as a soft loss penalty \cite{akhound2023}, and architecturally through
equivariant moving frames \cite{arora2024}.

This paper takes the architectural route for one specific classical object.
Ermakov \cite{ermakov1880} and Lewis \cite{lewis1967} showed that an auxiliary
function $\rho$ obeying the Pinney equation \cite{pinney1950}
\begin{equation}
  \ddot\rho + \omega^2(t)\,\rho = \frac{c}{\rho^3}
  \label{eq:pinney}
\end{equation}
generates an exact invariant for the parametric oscillator. We observe that the
same $\rho$ organises three superficially unrelated problems, that the sign of
$c$ is what distinguishes them, and that writing the associated change of
variables into the architecture costs no parameters. We call the resulting
model an \emph{Ermakov-based network} (\EB{}).

\paragraph{Contributions.}
\begin{enumerate}[leftmargin=*,itemsep=2pt]
\item \textbf{A moment identity that makes the embedding practical}
  (lemmas~\ref{lem:mom1} and~\ref{lem:mom2}): for the linear conservative
  families, $\rho$ follows in closed form from the initial data and the
  coefficients, with no auxiliary network and no knowledge of the solution.
\item \textbf{One embedding, three families}
  (sections~\ref{sec:famI}--\ref{sec:famIII}): the same $(\xi,\theta)$
  construction, differing only in the sign of $c$ and in which auxiliary factor
  --- amplitude or phase --- is attached.
\item \textbf{A comparative anatomy} (table~\ref{tab:synth}): each family needs
  a different component of the same construction, and we quantify each
  component's share.
\item \textbf{Ablation as a requirement, not a courtesy}
  (remark~\ref{rem:degenerate}): family~I is degenerate, in that a trivial time
  reparametrisation reproduces most of the reported gain and the whole of its
  transfer property. Any symmetry-informed comparison on such a problem is
  uninterpretable without that control. We regard this as the most portable
  lesson of the paper.
\item \textbf{A negative result we consider load-bearing}
  (remark~\ref{rem:nofixed}): extrapolation beyond the training horizon helps
  only when the reduced solution converges to a fixed point. When it beats
  instead, no model extrapolates, ours included.
\end{enumerate}

\paragraph{What we do not claim.} The scaling-variable reduction is classical
(remark~\ref{rem:classical}); so are the diffusive-time substitution and the
Lewis--Riesenfeld transformation. We claim none of them, nor a new architecture
class, nor superiority over neural operators, nor applicability beyond linear
first-kind self-similar problems.

\section{The Ermakov--Pinney backbone}
\label{sec:backbone}

\subsection{The invariant, for either sign of $c$}

Given $\ddot y+\omega^2(t)y=0$ and $\rho$ solving \eqref{eq:pinney}, the
Ermakov--Lewis quantity
\begin{equation}
  I=\tfrac12\Big[\,c\,(y/\rho)^2+(\rho\dot y-\dot\rho\,y)^2\Big]
  \label{eq:invariant}
\end{equation}
satisfies $\dot I=0$ identically. We stress \emph{for either sign of $c$}: the
familiar $c>0$ branch is oscillatory and underlies the Lewis--Riesenfeld theory
of the time-dependent quantum oscillator \cite{lewisriesenfeld1969}; the $c<0$
branch describes monotone expansion and is the one relevant to diffusion. Both
are used below. Pinney's formula gives the general solution of
\eqref{eq:pinney} as $\rho^2=Ay_1^2+2By_1y_2+Cy_2^2$ with $c=AC-B^2$ when the
Wronskian of $y_1,y_2$ is unity.

\subsection{The generic embedding}

All three families use the same three ingredients:
\begin{equation}
  \xi=\frac{x}{\rho(t)},\qquad
  \theta(t)=\int_0^t\frac{\kappa(s)}{\rho^2(s)}\,\rmd s,\qquad
  \text{field}=\rho^{-\alpha}\,\Lambda(x,t)\,\Phi(\xi,\theta),
  \label{eq:generic}
\end{equation}
where $\kappa$ is the diffusivity ($\kappa=D$) or $\hbar/m$ in the quantum
case, $\alpha$ is the scaling dimension fixed by the conserved quantity, and
$\Lambda$ is an auxiliary factor: $\Lambda\equiv1$ in the diffusive families
and $\Lambda=\rme^{\rmi S}$ with $S=m\dot\rho x^2/(2\hbar\rho)$ in the quantum
one. \emph{Which} of these ingredients carries the method turns out to depend
on the family, which is the comparative result of this paper.

\subsection{Three clocks}
\label{sec:clocks}

It pays to be explicit about what each candidate input measures time with.

\begin{center}\small
\begin{tabular}{llp{6.7cm}}
\toprule
clock & dimensions & what the solution does in it \\
\midrule
$t$ & time & evolves at a rate that varies with the coefficients \\
$\tau=\int_0^t\!D$ & length$^2$ & still spreads, but at a uniform rate:
   $m_2$ is linear in $\tau$ \\
$\theta=\int_0^t\!\kappa/\rho^2$ & dimensionless & does not spread:
   $\Phi(\cdot,\theta)$ is stationary or slowly varying \\
\bottomrule
\end{tabular}
\end{center}

\noindent The middle row invites a misreading that we have found easy to make.
The diffusive time $\tau$ is not ``time in units of the spreading''; it is time
in units of \emph{accumulated diffusivity}, with dimensions of length squared.
Only $\theta$ is the solution's own clock: $\rmd\theta$ measures elapsed time
against the time the solution currently needs to change by its own width, so
one unit of $\theta$ is one $e$-folding of $\rho$ wherever one stands on the
trajectory. That distinction is the content of the method, and it explains the
extrapolation results below: extrapolating in $\tau$ asks the network about
scales it never saw, whereas extrapolating in $\theta$ asks about a profile it
has already watched converge.

\section{Family I: linear diffusion ($c=-1$, $\omega\equiv0$)}
\label{sec:famI}

\begin{equation}
  u_t=D(t)\,u_{xx},\qquad u(\cdot,0)=u_0\ge0,\quad \textstyle\int u_0=1 .
  \label{eq:famI}
\end{equation}

\begin{lemma}[Second-moment identity]
\label{lem:mom1}
With enough decay, $m_2(t)=\int x^2u\,\rmd x$ satisfies $m_2(t)=m_2(0)+2\tau(t)$
exactly, where $\tau=\int_0^t D$.
\end{lemma}
\begin{proof}
$\frac{\rmd}{\rmd t}\int x^2u = D\!\int x^2u_{xx}
 = D[x^2u_x-2xu]_{-\infty}^{\infty}+2D\!\int u = 2D$.
\end{proof}

Setting $\rho=\sqrt{m_2(0)+2\tau}$, $\xi=x/\rho$, $\theta=\log(\rho/\rho_0)$
and $u=\rho^{-1}v$ gives, by a short computation (appendix~\ref{app:derA}),
\begin{equation}
  u_t-D u_{xx}=\frac{D}{\rho^3}\Big[v_\theta-\partial_\xi(v_\xi+\xi v)\Big],
  \label{eq:redI}
\end{equation}
so $v$ obeys the \emph{autonomous} Ornstein--Uhlenbeck equation whose unique
decaying unit-mass steady state is $v_\infty=(2\pi)^{-1/2}\rme^{-\xi^2/2}$. In
diffusive time $\rho$ solves $\rho''=-\rho^{-3}$, i.e.\ \eqref{eq:pinney} with
$\omega\equiv0$, $c=-1$.

\begin{remark}[The reduction is classical]
\label{rem:classical}
For constant $D$ this is the \emph{scaling-variable} change of the
dynamical-systems approach to self-similar asymptotics. Gallay and Wayne
\cite{gallaywayne2002} use $\xi=x/\sqrt{1+t}$, $\tau=\log(1+t)$ and
$\omega=(1+t)^{-1}w$, obtaining
$\partial_\tau w=\Delta_\xi w+\tfrac12(\xi\cdot\nabla_\xi)w+w$ with a Gaussian
kernel --- the same operator as \eqref{eq:redI} up to the factor $\tfrac12$, a
convention in the normalisation of $\rho$. The similarity ansatz is older still
\cite{barenblatt1996}, and $\tau=\int D$ is textbook \cite{crank1975}.
Lemma~\ref{lem:mom1} is what extends it to $D=D(t)$ and to unnormalised initial
data, and is therefore the part that makes an \emph{architectural} layer
possible.
\end{remark}

\begin{table}[t]\centering\small
\IfFileExists{tab_famI_en.tex}{\begin{tabular}{llcc}
\toprule
Model & Input & \multicolumn{2}{c}{Relative $L^2$ error} \\
\cmidrule(lr){3-4}
 & & Gaussian & bimodal \\
\midrule
PINN & $(x,t)$ & $3.84\times 10^{-3}$ {\scriptsize$\pm4.7\times 10^{-4}$} & $9.63\times 10^{-3}$ {\scriptsize$\pm2.6\times 10^{-3}$} \\
PINN-$\tau$ & $(x,\tau)$ & $1.76\times 10^{-3}$ {\scriptsize$\pm2.5\times 10^{-4}$} & $8.16\times 10^{-3}$ {\scriptsize$\pm2.7\times 10^{-3}$} \\
EB-coords & $(\xi,\theta)$ & $9.56\times 10^{-4}$ {\scriptsize$\pm1.4\times 10^{-4}$} & $6.33\times 10^{-3}$ {\scriptsize$\pm3.3\times 10^{-3}$} \\
\textbf{EB-Net} & $(\xi,\theta)$, $\rho^{-\alpha}v$ & $1.16\times 10^{-3}$ {\scriptsize$\pm4.7\times 10^{-4}$} & $5.53\times 10^{-3}$ {\scriptsize$\pm2.0\times 10^{-3}$} \\
\bottomrule
\end{tabular}
}{\textit{(missing)}}
\caption{\label{tab:famI}Family~I. Mean $\pm$ s.d.\ over five seeds; all models
have $\nparams$ parameters. The gain from PINN to PINN-$\tau$ uses no Ermakov
structure. The amplitude prefactor is neutral here (EB-coords vs \EB{}).}
\end{table}

\begin{remark}[This family is degenerate, and that is the lesson]
\label{rem:degenerate}
The map $t\mapsto\tau$ alone already removes $D(t)$ from \eqref{eq:famI}, and
the exact solution depends on $t$ \emph{only} through $\tau$. A model fed
$(x,\tau)$ therefore inherits both a large share of the accuracy gain and the
entire zero-shot transfer property, with no Ermakov structure whatsoever.
Table~\ref{tab:famI} shows PINN-$\tau$ recovering a factor $\FIagainTau$ of the
$\FIagainEB$ total. We conclude that family~I \emph{cannot} support a claim
about symmetry, and we present it not as evidence but as the reason the rest of
the paper is organised around controls.
\end{remark}

\subsection{Transfer across coefficient schedules certifies the trivial step}
\label{sec:transfer}

Because \eqref{eq:redI} contains no trace of $D$, the reduced profile is
universal across the family $\{D(\cdot)\}$: changing the coefficient only
changes the path $t\mapsto(\rho(t),\theta(t))$ traced in the $(\xi,\theta)$
plane. One is tempted to present the resulting zero-shot transfer as the
decisive argument for the embedding. Table~\ref{tab:transfer} shows why that
would be wrong.

\begin{table}[t]\centering\small
\IfFileExists{tab_transfer_en.tex}{\begin{tabular}{lccc}
\toprule
Evaluation $D(t)$ & PINN & PINN-$\tau$ & \textbf{EB-Net} \\
\midrule
$D_1$ (training) & $3.32\times 10^{-3}$ & $1.83\times 10^{-3}$ & $1.05\times 10^{-3}$ \\
$D_2$ (phase $+\pi/2$) & $6.12\times 10^{-2}$ & $1.83\times 10^{-3}$ & $8.84\times 10^{-4}$ \\
$D_3$ ($f{=}3$, $a{=}0.9$) & $5.35\times 10^{-2}$ & $1.85\times 10^{-3}$ & $1.01\times 10^{-3}$ \\
$D_4$ ($D_0{=}1.6$, fast) $\dagger$ & $1.26\times 10^{-1}$ & $1.14\times 10^{-2}$ & $1.81\times 10^{-3}$ \\
$D_5$ ($D_0{=}0.4$, slow) & $2.26\times 10^{-1}$ & $1.62\times 10^{-3}$ & $6.56\times 10^{-4}$ \\
\midrule
\multicolumn{4}{l}{\footnotesize $\dagger$ only $D_4$ leaves the trained diffusive-time interval.} \\
\bottomrule
\end{tabular}
}{\textit{(missing)}}
\caption{\label{tab:transfer}Zero-shot transfer in family~I: the same weights
evaluated under a different $D(t)$, with no retraining. PINN-$\tau$ matches
\EB{} on the schedules that stay inside the trained diffusive-time interval.}
\end{table}

\noindent On $D_2$, $D_3$ and $D_5$ --- which stay inside the trained
diffusive-time interval --- PINN-$\tau$ transfers as well as \EB{}, for the
reason given in remark~\ref{rem:degenerate}. The plain PINN has no mechanism at
all and degrades by one to two orders of magnitude. The informative column is
$D_4$, which overshoots the interval, together with the temporal extrapolation
to $t\in[T,2T]$: there \EB{} reaches $\extrapEbnet$ against $\extrapPinntau$
for PINN-$\tau$ and $\extrapPinn$ for the baseline. Two effects compound. The
logarithmic time contracts overshoot --- a $100\%$ excess in $\tau$ is only
$28.9\%$ in $\theta$ --- and the function being extrapolated has already
collapsed onto a constant profile.

\subsection{The physical residual carries an implicit $\rho^{-3}$ weighting}
\label{sec:weight}

Identity \eqref{eq:redI} has a consequence that any follow-up work in this
direction will have to account for. Minimising the \emph{physical} residual, as
a standard PINN does, is equivalent to minimising the reduced residual weighted
by $(D/\rho^3)^2$. Over our test horizon $\rho$ grows by a factor $3$, so early
times are weighted up to $27$ times more heavily than late ones. This bias
interacts with the amplitude ansatz: the model carrying the $\rho^{-\alpha}$
prefactor inherits a steeper weighting than the one that does not, which is why
the two are indistinguishable in table~\ref{tab:famI} even though only the
former makes the target exactly stationary.

\begin{table}[t]\centering\small
\IfFileExists{tab_weighted_en.tex}{\begin{tabular}{lcc}
\toprule
Model & physical residual & reduced residual \\
\midrule
PINN & $3.84\times 10^{-3}$ & $3.64\times 10^{-2}$ \\
PINN-$\tau$ & $1.76\times 10^{-3}$ & $1.99\times 10^{-2}$ \\
EB-coords & $9.56\times 10^{-4}$ & $1.05\times 10^{-2}$ \\
\textbf{EB-Net} & $1.16\times 10^{-3}$ & $1.04\times 10^{-2}$ \\
\bottomrule
\end{tabular}
}{\textit{(missing)}}
\caption{\label{tab:weighted}Family~I with the residual reweighted by
$\rho^3/D$ (normalised to unit mean), applied identically to all four models.
Absolute accuracy worsens by about an order of magnitude; the ordering
survives.}
\end{table}

\noindent Reweighting makes every model roughly an order of magnitude worse in
physical $L^2$ ($\FIaPinn\to\WmeanPinn$ for the baseline): the implicit bias is
not a defect for this metric, since early times are where the profile is narrow
and hardest to resolve. The result we care about is that the \emph{ordering}
survives ($\WmeanPinn > \WmeanPinntau > \WmeanEbc \approx \WmeanEbnet$). Since
the ablation ladder is invariant under this rather drastic change of loss
weighting, the gain is attributable to the coordinates rather than to a
fortunate weighting.

\section{Family II: diffusion in a breathing trap ($c=-1$, $\omega\neq0$)}
\label{sec:famII}

\begin{equation}
  u_t=D(t)u_{xx}+\gamma(t)\,(xu)_x,\qquad \gamma>0 .
  \label{eq:famII}
\end{equation}

The drift contains $x$ explicitly, so no reparametrisation $s=s(t)$ can remove
it: the degeneracy of family~I is gone. Physically this is diffusion inside a
trap whose stiffness is modulated --- optical tweezers at varying power, a
colloid in an oscillating field.

\begin{lemma}[Moment identity, generalised]
\label{lem:mom2}
$\rmd m_2/\rmd t=2D(t)-2\gamma(t)m_2$, a linear ODE; hence $\rho=\sqrt{m_2}$ is
again available in closed form from $m_2(0)$ alone.
\end{lemma}

Writing $u=\rho^{-1}v(\xi,\theta)$ with $\xi=x/\rho$ and $\theta'=D/\rho^2$
reproduces \emph{exactly} the operator of \eqref{eq:redI} provided
$\rho\dot\rho=D-\gamma\rho^2$; differentiating that relation gives
\begin{equation}
  \ddot\rho+(\dot\gamma-\gamma^2)\rho-\frac{\dot D}{\rho}=-\frac{D^2}{\rho^3},
  \label{eq:pinneyII}
\end{equation}
which for constant $D$ is \eqref{eq:pinney} with $\omega^2=\dot\gamma-\gamma^2$
and $c=-D^2$. Two consequences matter. First, $\omega\neq0$: the Ermakov
structure is no longer in its degenerate branch. Second, $\rho$
\emph{oscillates} rather than growing as a power of $t$, so the classical
scaling variables $\xi=x/\sqrt{t}$, $\theta=\log t$ of
remark~\ref{rem:classical} do not describe this case at all.

\begin{table}[t]\centering\small
\IfFileExists{tab_famII_en.tex}{\begin{tabular}{lcccc}
\toprule
Model & \multicolumn{2}{c}{Relative $L^2$ error} & \multicolumn{2}{c}{Extrapolation $t\in[T,2T]$} \\
\cmidrule(lr){2-3}\cmidrule(lr){4-5}
 & Gaussian & bimodal & Gaussian & bimodal \\
\midrule
PINN & $1.19\times 10^{-2}$ {\scriptsize$\pm1.2\times 10^{-3}$} & $1.74\times 10^{-2}$ {\scriptsize$\pm1.4\times 10^{-3}$} & $1.21\times 10^{0}$ & $1.14\times 10^{0}$ \\
PINN-$\tau$ & $8.00\times 10^{-2}$ {\scriptsize$\pm4.5\times 10^{-2}$} & $1.13\times 10^{-1}$ {\scriptsize$\pm3.6\times 10^{-2}$} & $4.16\times 10^{-1}$ & $2.89\times 10^{-1}$ \\
EB-coords & $4.13\times 10^{-2}$ {\scriptsize$\pm4.9\times 10^{-2}$} & $9.80\times 10^{-2}$ {\scriptsize$\pm2.0\times 10^{-2}$} & $6.31\times 10^{-1}$ & $2.86\times 10^{0}$ \\
\textbf{EB-Net} & $1.01\times 10^{-3}$ {\scriptsize$\pm9.7\times 10^{-5}$} & $6.32\times 10^{-3}$ {\scriptsize$\pm6.3\times 10^{-4}$} & $6.81\times 10^{-3}$ & $7.09\times 10^{-2}$ \\
\bottomrule
\end{tabular}
}{\textit{(missing)}}
\caption{\label{tab:famII}Family~II. Mean $\pm$ s.d.\ over three seeds.
PINN-$\tau$ is now \emph{worse} than the plain baseline; the amplitude
prefactor separates EB-coords from \EB{} by $\FIIagainCoords\times$.}
\end{table}

\begin{figure}[t]\centering
\includegraphics[width=\linewidth]{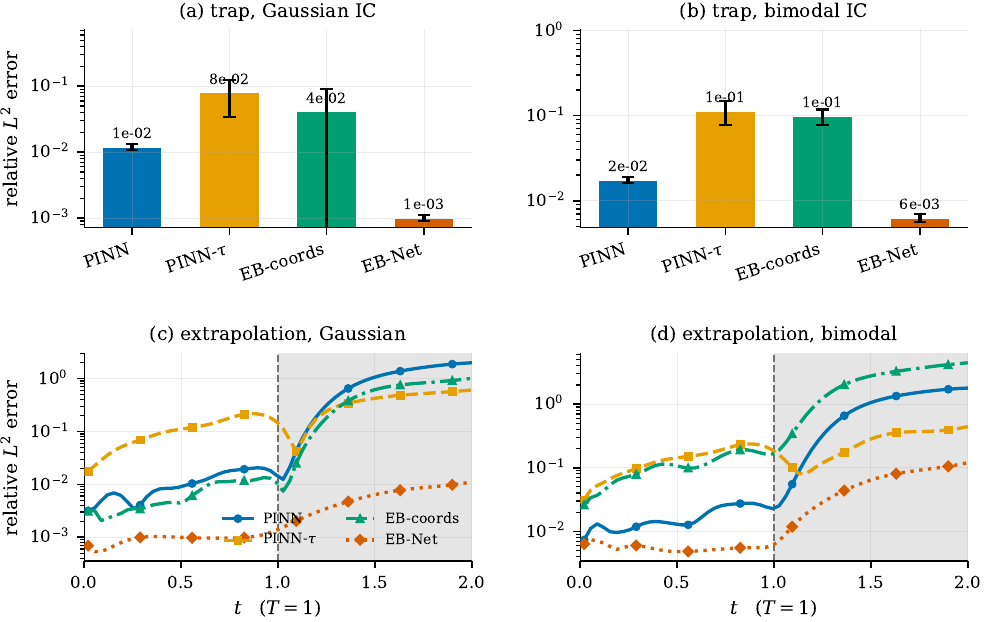}
\caption{\label{fig:famII}Family~II. (a,b) in-domain error; (c,d)
instantaneous error for weights trained only on $t\in[0,T]$, shaded region
extrapolated.}
\end{figure}

\paragraph{The trivial control does not merely fail; it hurts.} PINN-$\tau$
reaches $\FIIaPinntau$ against the baseline's $\FIIaPinn$, with a seed-to-seed
spread an order of magnitude larger than any other model's. Feeding $\tau$
supplies a clock that absorbs $D$ but is inconsistent with the drift, and the
network is worse off than with physical time. We regard this as the cleanest
available refutation of the objection that family~I invites.

\paragraph{The amplitude prefactor switches from neutral to essential.} It was
indistinguishable from omission in family~I; here removing it costs
$\FIIagainCoords\times$ (Gaussian) and $\FIIbgainCoords\times$ (bimodal), and
in the bimodal case EB-coords is worse than the plain PINN. The trap compresses
and expands the density, so the amplitude genuinely oscillates and the ansatz
must absorb it.

\paragraph{Extrapolation.} With weights trained only on $[0,T]$, \EB{} reaches
$\FIIaEbnetex$ on $t\in[T,2T]$ against the baseline's $\FIIaPinnex$, a factor
$\FIIaexRatio$. Figure~\ref{fig:famII}(c) shows \EB{} crossing $t=T$ without
visible incident.

\section{Family III: the quantum oscillator ($c=+1$)}
\label{sec:famIII}

With $\hbar=m=1$,
\begin{equation}
  \rmi\psi_t=-\tfrac12\psi_{xx}+\tfrac12\omega^2(t)x^2\psi .
  \label{eq:famIII}
\end{equation}
The Lewis--Riesenfeld transformation \cite{lewisriesenfeld1969}
\begin{equation}
  \psi=\rho^{-1/2}\exp\!\Big(\frac{\rmi\dot\rho x^2}{2\rho}\Big)\varphi(\xi,\theta),
  \qquad \xi=\frac{x}{\rho},\quad \dot\theta=\frac{1}{\rho^2},
  \label{eq:LR}
\end{equation}
with $\rho$ solving \eqref{eq:pinney} at $c=+1$, yields the \emph{autonomous}
oscillator $\rmi\varphi_\theta=-\tfrac12\varphi_{\xi\xi}+\tfrac12\xi^2\varphi$.
We verified this symbolically; the residual difference is exactly zero
(appendix~\ref{app:code}). Here $\alpha=1/2$ because $|\psi|^2$ is conserved,
and $\Lambda=\rme^{\rmi S}$ is not optional --- omitting it breaks the
reduction, a claim we measure rather than assert.

\paragraph{The honest control is adiabatic, not trivial.} No time
reparametrisation absorbs $\omega(t)$. The baseline a physicist would reach for
is the adiabatic one: rescale by the instantaneous oscillator length
$1/\sqrt{\omega}$ and use the adiabatic phase $\int\omega\,\rmd t$. It is exact
in the slow limit. We therefore choose a regime where it has no right to work:
$\omega\in[0.3,1.7]$, $\rho$ breathing by a factor $2.4$, and
$|\dot\omega|/\omega^2=2.2$.

\paragraph{The invariant does physical work.} Evaluating \eqref{eq:invariant}
as a quantum expectation on the exact solution gives
$\langle I\rangle=0.500000000000$ at every time to $2.2\times10^{-16}$, while
$\langle H\rangle$ runs from $0.500$ to $0.960$. The energy is not conserved;
the Ermakov--Lewis invariant is, and its value is exactly the ground-state
energy of the reduced oscillator.

\begin{table}[t]\centering\small
\IfFileExists{tab_famIII_en.tex}{\begin{tabular}{llcc}
\toprule
Model & Input & \multicolumn{2}{c}{Relative $L^2$ error} \\
\cmidrule(lr){3-4}
 & & ground state & superposition \\
\midrule
PINN & $(x,t)$ & $9.99\times 10^{-2}$ {\scriptsize$\pm6.1\times 10^{-3}$} & $3.40\times 10^{-1}$ {\scriptsize$\pm9.5\times 10^{-2}$} \\
PINN-ad & $(x\sqrt{\omega},\int\!\omega)$ & $1.16\times 10^{-1}$ {\scriptsize$\pm2.1\times 10^{-2}$} & $4.19\times 10^{-1}$ {\scriptsize$\pm9.5\times 10^{-2}$} \\
EB-coords & $(\xi,\theta)$ & $4.10\times 10^{-2}$ {\scriptsize$\pm3.2\times 10^{-3}$} & $1.14\times 10^{-1}$ {\scriptsize$\pm1.7\times 10^{-2}$} \\
EB no-phase & $(\xi,\theta)$, $\rho^{-1/2}\varphi$ & $4.19\times 10^{-2}$ {\scriptsize$\pm3.9\times 10^{-3}$} & $9.16\times 10^{-2}$ {\scriptsize$\pm2.7\times 10^{-2}$} \\
\textbf{EB-Net} & $(\xi,\theta)$, $\rho^{-1/2}e^{iS}\varphi$ & $7.31\times 10^{-3}$ {\scriptsize$\pm1.0\times 10^{-3}$} & $6.07\times 10^{-2}$ {\scriptsize$\pm1.6\times 10^{-2}$} \\
\bottomrule
\end{tabular}
}{\textit{(missing)}}
\caption{\label{tab:famIII}Family~III. Mean $\pm$ s.d.\ over three seeds. The
adiabatic control is worse than the plain PINN in both cases. Omitting the
quadratic phase costs $\FIIIagainPhase\times$ on the ground state.}
\end{table}

\begin{figure}[t]\centering
\includegraphics[width=\linewidth]{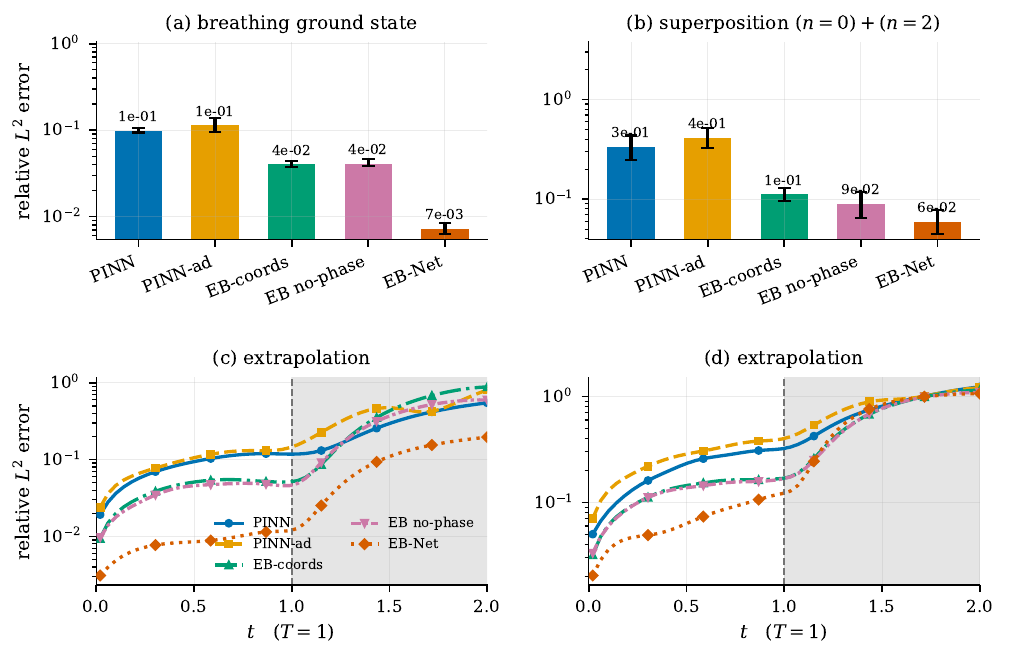}
\caption{\label{fig:famIII}Family~III. Note in (b,d) that no model extrapolates
when the reduced solution beats rather than converging
(remark~\ref{rem:nofixed}).}
\end{figure}

\begin{remark}[When the method does not help]
\label{rem:nofixed}
For the $(n{=}0)+(n{=}2)$ superposition the reduced solution is not stationary:
it beats at $E_2-E_0=2$ in $\theta$. There is then no fixed point to
extrapolate toward, and figure~\ref{fig:famIII}(d) shows every model, ours
included, degrading to $\sim0.8$ past $t=T$. The extrapolation advantage of the
embedding is therefore \emph{conditional on convergence of the reduced
solution}, not a general property. We prefer this sharp, falsifiable statement
to a blanket claim.
\end{remark}

\section{Synthesis: which component carries the method}
\label{sec:synth}

\begin{table}[h]\centering\small
\IfFileExists{tab_synth_en.tex}{\begin{tabular}{llllcc}
\toprule
Family & Pinney branch & Trivial control & Load-bearing piece & vs PINN & piece's share \\
\midrule
I. diffusion $u_t=Du_{xx}$ & $c=-1$, $\omega=0$ & $\tau$ (suffices) & coordinates only & 4.0 & prefactor: 1.22$\times$ \\
II. trap $+\gamma(xu)_x$ & $c=-1$, $\omega\neq0$ & $\tau$ (fails) & $+$ prefactor $\rho^{-\alpha}$ & 11.7 & prefactor: 41$\times$ \\
III. quantum $\omega(t)$ & $c=+1$ & adiabatic (fails) & $+$ phase $e^{iS}$ & 13.7 & phase: 5.7$\times$ \\
\bottomrule
\end{tabular}
}{\textit{(missing)}}
\caption{\label{tab:synth}The central result. One construction, three families,
and a different component load-bearing in each. Gains are \EB{} versus the
plain PINN on the favourable initial condition of each family; the last column
isolates the named component.}
\end{table}

\noindent Read across the rows, table~\ref{tab:synth} says three things. The
\emph{control} degrades from sufficient to actively harmful as one moves away
from the degenerate branch. The \emph{gain} over the plain baseline grows in
the same direction. And the \emph{component} that matters changes: bare
coordinates, then amplitude, then phase. A reader who saw only family~I would
conclude the prefactor is decorative; one who saw only family~III would
conclude the same of the amplitude. Both conclusions would be artefacts of the
chosen test problem, which is our main methodological point.

\section{Architecture and protocol}
\label{sec:protocol}

\EB{} has three blocks: a parameter-free embedding $(x,t)\mapsto(\xi,\theta)$;
an MLP backbone; and a reconstruction $\rho^{-\alpha}\Lambda\,\Phi$. Because
the embedding carries no weights, every model in every ladder has an identical
parameter count ($\nparams$ for the real families) and the same per-iteration
cost to within $4\,\%$.

\paragraph{Fairness.} Within each family all models (i) minimise the
\emph{same} loss functional, evaluated by automatic differentiation in
\emph{physical} coordinates --- no model is trained on the reduced equation;
(ii) share architecture ($4\times64$, $\tanh$, Xavier initialisation),
collocation budget and optimiser schedule (Adam \cite{kingmaba2015} with cosine
decay, then L-BFGS); (iii) run over the same seeds; and (iv) normalise inputs
to $O(1)$ in their own natural parametrisation. The initial-condition weight
was selected per model by a sweep on a held-out seed, which selected the same
value ($\lambda_{\mathrm{ic}}=10$) for all of them.

\paragraph{References and precision.} The scale functions are obtained by
integrating the relevant ODE system to \texttt{rtol}$=10^{-12}$ and exposed to
automatic differentiation through a $C^1$ cubic Hermite interpolant built from
exact nodal derivatives. Exact solutions satisfy their PDEs to $10^{-10}$ or
better on the evaluation grid, so the reported errors are not
reference-limited.

\section{Related work}
\label{sec:related}

\paragraph{Scaling variables.} The reduction of section~\ref{sec:famI} is
classical \cite{gallaywayne2002,barenblatt1996}, as is the diffusive-time
substitution \cite{crank1975} and the Lewis--Riesenfeld transformation
\cite{lewisriesenfeld1969}. Our role is to move them from the analysis into the
architecture, which lemmas~\ref{lem:mom1}--\ref{lem:mom2} make possible for
time-dependent coefficients.

\paragraph{Symmetry in neural solvers.} Lie point symmetries have been used for
augmentation \cite{brandstetter2022}, as a loss penalty \cite{akhound2023}, and
structurally for ODEs \cite{arora2024}; conservative projections appear in
\cite{cardosobihlo2023}. Transformations of PINN inputs to smooth the target
\cite{vtpinn2024} and architectures for variable-coefficient PDEs
\cite{miao2023vcpinn} are the nearest PINN-side neighbours.

\paragraph{Dynamic rescaling.} Self-similar blow-up profiles have been found by
fixing a similarity ansatz \cite{wang2023euler}. The closest work is
Kavousanakis \etal \cite{kavousanakis2025}, who \emph{learn} the rescaling with
an auxiliary network under algebraic constraints, producing an index-2
differential-algebraic system. That method applies when the scaling is unknown,
including second-kind similarity; ours applies when it is known, which by our
lemmas is the whole linear conservative class, and costs no extra parameters.

\section{Limitations}
\label{sec:limits}

\begin{itemize}[leftmargin=*,itemsep=2pt]
\item \textbf{Linearity and conservation.} Both lemmas use them; for nonlinear
  problems (porous medium, Fisher--KPP) the moment hierarchy does not close and
  $\rho$ is no longer explicit.
\item \textbf{First-kind similarity only.} Second-kind exponents, fixed by a
  nonlinear eigenvalue problem \cite{barenblatt1996}, are outside the
  construction.
\item \textbf{Conditional extrapolation.} Remark~\ref{rem:nofixed}.
\item \textbf{Operator-learning baselines are a different comparison.} A neural
  operator \cite{li2021fno,lu2021deeponet} trained across the coefficient
  family would be the natural rival for the transfer claim of
  section~\ref{sec:transfer}, and we have not run it. We note that the
  comparison is not like-for-like: an operator network amortises training over
  a distribution of coefficients, whereas every model compared here solves a
  single instance from its residual alone, with no data. The operator setting
  is the subject of the companion $N$-dimensional work
  (section~\ref{sec:outlook}), where it is the main object rather than a
  baseline.
\item \textbf{Absolute accuracy in family~III.} All models remain near
  $10^{-1}$ on the superposition at our training budget; the comparison is
  informative but the absolute numbers are not competitive with a spectral
  solver.
\item \textbf{Scope.} One spatial dimension throughout.
\end{itemize}

\section{Outlook: the $N$-dimensional case}
\label{sec:outlook}

The one-dimensional construction generalises, but not by replacing scalars with
matrices. For a general symmetric frequency matrix $\Omega^2(t)$ whose
eigenvectors rotate, the correct frame follows from the \emph{linear} equation
$\ddot Q+\Omega^2Q=0$ rather than from a symmetric matrix Pinney equation,
which fails to preserve symmetry as soon as $[\Omega^2(t),\Omega^2(t')]\neq0$;
the width and phase matrices then obey a matrix Ermakov--Pinney system closed
by a Lyapunov equation. The scalar Ermakov clock $\theta$ becomes a
$U(N)$-valued clock carrying a rotational term, so no single latent time exists
unless the frame is isotropic, and the $N$ Williamson invariants generalise
\eqref{eq:invariant}. The frame remains observable from phase-space moments,
which additionally permits reconstruction of an unknown $\Omega^2(t)$. These
results, their operator-learning consequences and their verification against a
spectral solution of the two-dimensional Schr\"odinger equation with a chirped,
cross-coupled trap are developed in a companion paper in preparation. The
present work establishes the one-dimensional case that the companion assumes.

\section{Conclusion}

A single Pinney function organises linear diffusion, diffusion in a breathing
trap, and the quantum oscillator with time-dependent frequency; the sign of $c$
is what separates them. Writing the associated change of variables into a
neural solver's input layer costs no parameters and is made practical for
time-dependent coefficients by an exact moment identity. What the construction
buys depends on the family in a way we have quantified: coordinates alone where
a time reparametrisation would have sufficed, the amplitude prefactor where a
trap makes the density oscillate, the quadratic phase where the dynamics are
quantum. The honest control degrades from sufficient to harmful along the same
axis. We regard the resulting table, rather than any single error figure, as
the contribution.

\section*{Acknowledgements}
The authors thank the transformative agreements of Universidad
Aut\'onoma Metropolitana.

\section*{Data availability statement}
No new data were created in this study beyond those generated by the
accompanying code. All results, figures and tables are reproduced from
scratch by the software archived at \texttt{[DOI on acceptance]}; no
numerical value in this paper is transcribed by hand.

\section*{Declaration of generative AI use}
During the preparation of this work the authors used a large language
model to assist with code implementation, numerical experimentation and
manuscript drafting. The authors reviewed and edited all content and
take full responsibility for the published work.

\section*{Conflicts of interest}
The authors declare no conflicts of interest.

\appendix

\section{Derivation for family I}
\label{app:derA}

In diffusive time $s=\tau$, \eqref{eq:famI} reads $u_s=u_{xx}$ and
$\rho\rho'=1$ since $\rho^2=m_2(0)+2s$. With $u=\rho^{-1}v(\xi,\theta)$,
$\xi=x/\rho$, $\theta'=\rho^{-2}$:
\begin{equation}
  u_s =-\rho^{-2}\rho'v-\rho^{-2}\rho'\xi v_\xi+\rho^{-1}\theta'v_\theta,
  \qquad u_{xx}=\rho^{-3}v_{\xi\xi}.
\end{equation}
Multiplying $u_s-u_{xx}$ by $\rho^3$ and using $\rho\rho'=\rho^2\theta'=1$
gives $v_\theta-v_{\xi\xi}-\xi v_\xi-v$; multiplying by $D/\rho^3$ and using
$u_t=D u_s$ gives \eqref{eq:redI}. Integrating $\theta'=\rho^{-2}$ yields
$\theta=\log(\rho/\rho_0)$. The exponent $\alpha=1$ is fixed by mass
conservation: $\int u\,\rmd x=\rho^{1-\alpha}\int v\,\rmd\xi$ is
$t$-independent only for $\alpha=1$; in $d$ dimensions $\alpha=d$, and in the
quantum case $\alpha=1/2$ because $\int|\psi|^2$ is the conserved quantity.

\section{Code}
\label{app:code}

Implementation: \texttt{ebnet.py} (core and family~I), \texttt{trap.py} and
\texttt{run\_trap.py} (family~II), \texttt{quantum.py} and
\texttt{run\_quantum.py} (family~III), plus \texttt{verify\_*.py} for the
symbolic and numerical checks and \texttt{make\_tables\_final.py}, which
regenerates every number in this paper from the stored results.

\begin{lstlisting}[caption={Family II: the scale function comes from a linear
moment ODE, so no auxiliary network is needed.}]
def _build(self, n=6001):
    def rhs(t, y):
        m2, _ = y
        return [2*self._Dnp(t) - 2*self._gnp(t)*m2,   # Lemma 2
                self._Dnp(t) / m2]                    # dtheta/dt
    ts  = np.linspace(0.0, self.t_end, n)
    sol = solve_ivp(rhs, (0.0, self.t_end), [self.m2_0, 0.0], t_eval=ts,
                    rtol=1e-12, atol=1e-14, method="DOP853")
    m2, th = sol.y
    rho  = np.sqrt(m2)
    drho = (self._Dnp(ts) - self._gnp(ts)*m2) / rho   # exact nodal derivative
    self._rho   = Hermite(ts, rho, drho)              # C^1, differentiable
    self._theta = Hermite(ts, th, self._Dnp(ts)/m2)
\end{lstlisting}

\begin{lstlisting}[caption={Family III: the full Lewis--Riesenfeld
reconstruction. The ablated model drops cS/sS, which is the phase-factor
comparison of table~\ref{tab:famIII}.}]
class QEBNet(_QSolver):
    def forward(self, x, t):
        rho, drho = self.p.rho(t), self.p.drho(t)
        f  = self.mlp(torch.cat([x/rho, self.p.theta(t)/self._thmax], 1))
        fr, fi = f[:, :1], f[:, 1:]
        S  = drho * x**2 / (2*rho)              # quadratic phase
        cS, sS = torch.cos(S), torch.sin(S)
        pre = rho**(-0.5)
        return torch.cat([pre*(fr*cS - fi*sS),
                          pre*(fr*sS + fi*cS)], dim=1)
\end{lstlisting}

\begin{lstlisting}[caption={Symbolic verification of the Lewis--Riesenfeld
transformation: the printed difference is exactly zero.}]
psi  = rho**sp.Rational(-1,2) * sp.exp(I*rp*x**2/(2*rho)) * phi(x/rho, th)
expr = I*sp.diff(psi,t) + sp.Rational(1,2)*sp.diff(psi,x,2) \
       - sp.Rational(1,2)*w**2*x**2*psi
expr = expr.subs(sp.Derivative(rho,(t,2)), 1/rho**3 - w**2*rho).doit()
expr = expr.subs(sp.Derivative(th,t), 1/rho**2)
# -> i phi_theta + 1/2 phi_xixi - 1/2 xi^2 phi        (difference = 0)
\end{lstlisting}

\bibliographystyle{unsrt}
\bibliography{refs}
\end{document}